\documentclass[letterpaper, 10 pt, conference]{ieeeconf}  %
\IEEEoverridecommandlockouts                              %
\usepackage[T1]{fontenc} 
\usepackage{enumerate}
\usepackage{amsfonts}
\usepackage[colorinlistoftodos]{todonotes}
\usepackage[font=small,skip=12pt]{caption}
\usepackage{float}
\usepackage{subcaption}
\usepackage[ruled,vlined]{algorithm2e}
\makeatletter
\let\NAT@parse\undefined
\makeatother
\usepackage{hyperref}
\usepackage{verbatim}
\usepackage{amsmath}
\title{\LARGE \bf Auto-calibration Method Using Stop Signs for Urban Autonomous Driving Applications

}

\author{$^{*}$Yunhai Han$^{1}$, $^{*}$Yuhan Liu$^{1}$, David Paz$^{1}$, Henrik Christensen$^{1}$
\thanks{*Equal contributions. }
\thanks{$^{1}$Contextual Robotics Institute, University of California, San Diego (UCSD), 9500 Gilman Dr, La Jolla, CA, 92093}}%

\begin{document}
\maketitle
\thispagestyle{empty}
\pagestyle{empty}

\begin{abstract}

Calibration of sensors is fundamental to robust performance for intelligent vehicles. In natural environments, disturbances can easily challenge calibration. One possibility is to use natural objects of known shape to recalibrate sensors. An approach based on recognition of traffic signs, such as stop signs, and use of them for recalibration of cameras is presented. The approach is based on detection, geometry estimation, calibration, and recursive updating. Results from natural environments are presented that clearly show convergence and improved performance.

\end{abstract}

\section{INTRODUCTION}
Advanced driver assistance systems (ADAS) and partial autonomy is emerging from numerous companies. Robustness in performance is key to success. The most promising sensor modality is imaging, due to low-cost and richness of potential information to be computed from images. 
A camera system is frequently used in distance perception, speed inference, and object detection.
For such applications, the camera 
parameters are often essential to reconstruct parts of a scene and influence the accuracy of
measurements.

One can calibrate a camera once before the beginning of a 
mission that requires minimal camera motion, as the intrinsic parameters are 
assumed to be static.
However, when the camera is mounted on a vehicle, the camera intrinsic parameters
vary due to the mechanical vibrations caused by bumps in the road,
or temperature fluctuations caused by the weather \cite{Dang2009ContinuousSS,labe2004geometric,smith2012effects}. It is also a challenge to repeatedly calibrate a camera for long-term use.   
Long-term operations and the maintenance-free advantages of a camera system can only be guaranteed by auto-calibration methods, which can update the camera parameters as needed.

The commonly used camera calibration methods (e.g. \cite{Heik1997calib}) require some special reference patterns, a chessboard, for example, whose physical size is known beforehand. However, these special patterns are not readily available when the vehicle is on the road. In this case, the auto-calibration system is necessary to extract the reference information from its environment. 


\begin{figure}[t]
\centering
\includegraphics[angle=0,height=6cm,width=7.4cm]{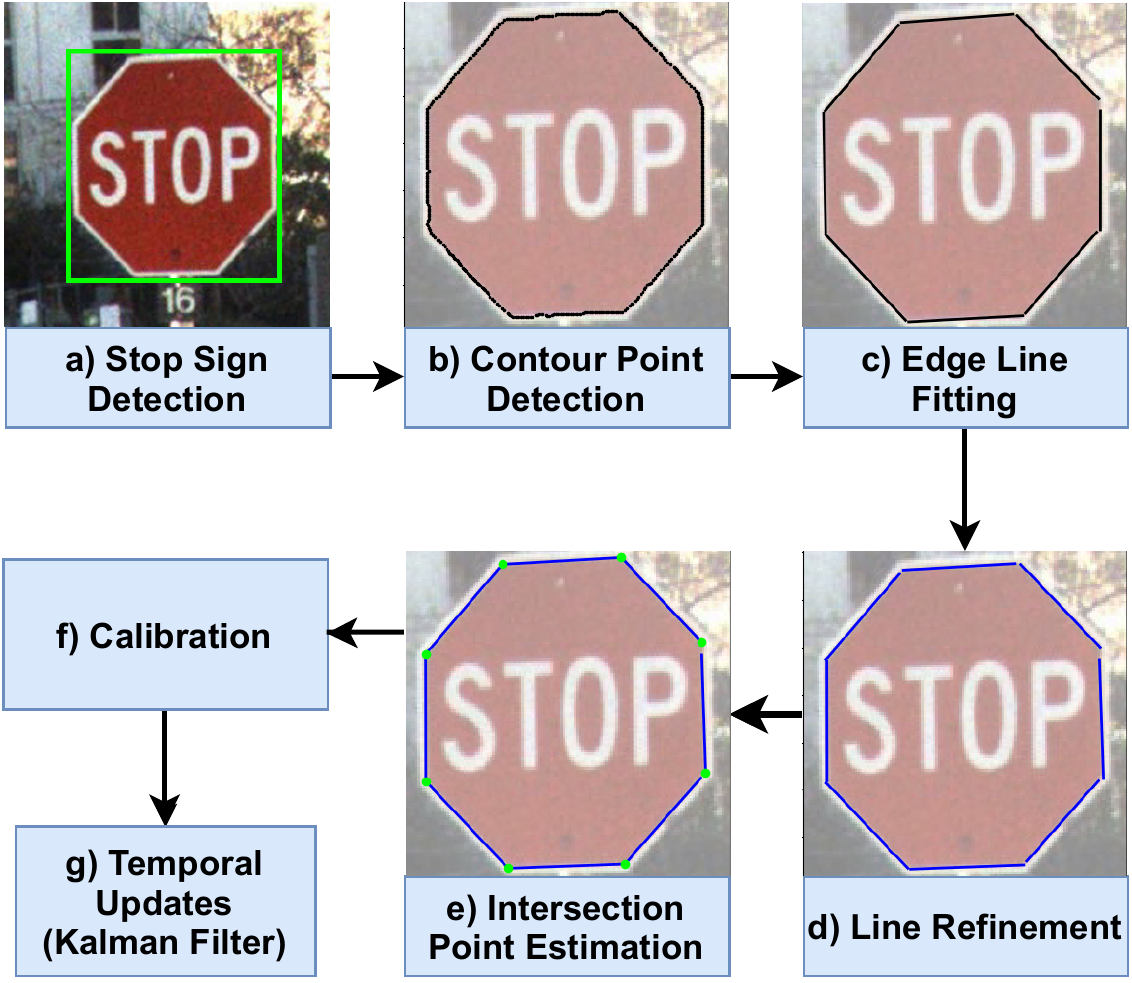}
\caption{A demonstration of the system framework. (a) The bounding box for the stop sign detected by Mask R-CNN. (b) The contour points (black dots) of the inner octagon. (c)-(d) The estimated lines/edges. The black solid lines and the blue solid lines represent the estimated lines before and after line refinement, respectively. (e). The intersection/corner points. The green circles are the estimated intersections. The eight corner points are then used for homography estimation together with reference information.} 
\label{fig:output}
\end{figure}

A few auto/self-calibration methods has been proposed. Ribeiro \cite{ribeiro2006automatic}, Hai \cite{hai2014novel}, and Lu \cite{Lu2014calib} extracts feature points from the road lines for calibration purpose, yet it is impossible to assume the same width of the road lines in different regions. Also, because of wear, imperfections exist along the contours of road lines. These artifacts and variations can lead to calibration errors. In \cite{broggi2001selfcalibration}, two solutions are presented. One designs a structured environment (a grid of known size painted on the road) that the cameras detect as the vehicle drives near. The other method uses several markers placed on the hood of the car. However, this can distract other drivers, assumes that the hood is within the FOV of the camera, and the performance can vary across different vehicles if the hood is not flat. Both are not suitable for real driving scenarios.

Stop signs, unlike a designed pattern, are ubiquitous. The physical size of stop signs is standardized by the U.S. Department of Transportation \cite{mutcd2003}. 

Compared with other reference information existing in the urban environments, there are three advantages of using traffic signs, and stop signs in particular:
\begin{itemize}
    \item Adequate feature points that can be detected with sub-pixel accuracy (at least eight from the inner octagon corners of the stop sign).
    \item Compared with other reference objects (e.g., road lines), the relative geometric properties of a stop sign remain unchanged, because they are made of metals which are robust to external forces.
    \item A vehicle slows down when approaching a stop sign, reducing the image blur and the rolling shutter effects.
\end{itemize}
Therefore, we present an approach to auto-calibrate with stop signs. The rest of the paper is organized as follows: Section \ref{section: related work} gives a brief summary of the related work; Section \ref{section: methodology} presents the details of our approach; Section \ref{section: experiments} shows the experiment results on the data collected from the UCSD campus using four vehicular cameras.

\section{RELATED WORK} \label{section: related work}
\textbf{Camera Calibration:}
Camera parameters can be estimated from enough pairs of 3D object points and corresponding 2D image points \cite{hartley2004multiple}. In \cite{zhang2000calibration}, Zhang proposed to calibrate a camera with a planar object, opening the door to the famous chessboard calibration method \cite{Kaehler2016LearningO3}.

\textbf{Auto-calibration:}
Several researchers have contributed to the auto-calibration using different reference objects. In \cite{ribeiro2006automatic,hai2014novel,Lu2014calib}, vanishing points or road corners were extracted from the dotted road lines for online calibration. In \cite{bhardwaj2018autocalib}, predefined feature points from other automobiles were used to build the correspondence model, applied to obtain traffic camera parameters. In \cite{dubska2014automatic}, surveillance camera calibration was based on three vanishing points defining the stream of vehicles.
In \cite{lv2006camera}, the same work was done through representing a walking human as a vertical line segment. Besides, \cite{broggi2001selfcalibration} placed markers on the vehicle's hood, from which the relative positions and orientations of the on-board cameras can be estimated quickly.

\textbf{Traffic Sign Detection:}
Object detection has been a well-studied area with the help of convolutional neural networks (CNNs). Proposal-free architectures (e.g. YOLO \cite{redmon2015yolo}, YOLOv3 \cite{redmon2018yolo}) can easily run in real-time (45 Hz), while proposal-based ones (e.g. Faster R-CNN \cite{ren2015fasterrcnn}, FPN \cite{lin2016fpn}), though requiring more resources and typically taking longer to run, generally outperform the former in terms of accuracy, especially with small objects and complicated scenes. The instance segmentation method in \cite{he2017maskrcnn}, which labels the object pixels, can also be applied to find the stop sign bounding boxes. Besides, optimized architectures (e.g. \cite{wu2013traffic}, \cite{zhu2016traffic}) tailored for the traffic sign detection have been proposed as well. 

\textbf{Shape Matching:}
Shape matching checks the resemblance of two shapes using a similarity measure \cite{veltkamp2001shape}. Particularly in our case, the shape enclosed by the detected stop sign edges has to be an octagon. Arkin et al. \cite{arkin1991polygon} represented the polygons with the turning functions and measured the $L_p$ distances. 
Huttenlocher and Kedem \cite{huttenlocher1990hausdorff} moved further by considering (in addition to rotation, translation, and scale invariance) the affine invariance.

\section{METHODOLOGY} \label{section: methodology}
The automatic system can be divided into seven modules (see Fig. \ref{fig:output}).
The first five parts generate the 2D-3D correspondence pairs of corner points from an urban image sequence, followed by the last two steps calculating the intrinsic parameters.

\subsection{Stop Sign Detection} 
A Mask R-CNN model with a ResNet-50 and Feature Pyramid Network backbone \cite{wu2019detectron2} were used to generate 2D bounding boxes on the image frames (see Fig. \ref{fig:output}(a)). The network was pre-trained on COCO2017 \cite{lin2014coco}.

Compared to Faster R-CNN, Mask R-CNN has a higher stop sign detection rate, which potentially increases our system's calibration accuracy by providing more reference candidates. The relatively low speed of a Mask R-CNN model doesn't affect our current non-real time system, because the overhead lies mostly in the Edge Line Fitting module (section \ref{subsection: edge}).

\subsection{Contour Points Detection through Color Segmentation} \label{subsection: contour}
This module detects the contour points of the red inner octagon from a stop sign (Fig. \ref{fig:output}(b)). An image pre-processing of color space conversion (from RGB to HSV) proves to be beneficial for the red color extraction with a threshold. Afterwards, 
on the HSV images,
the Canny/Devernay algorithm is used to attain the contour points of the octagon with sub-pixel accuracy:
\begin{itemize}
    \item [a.] The Canny Algorithm
    
     Canny \cite{canny1986edge} applies the first derivative of a Gaussian filter over an image to get the image gradient, of which the edge points are the local maxima along the normal direction $\mathbf{n}$. The direction $\mathbf{n}$ orthogonal to an edge can be approximated with the image gradient:
    \begin{equation}
     \mathbf{n} \approx \frac{\nabla(G \star I)}{|\nabla(G \star I)|},
    \end{equation}
    where $G$ and $I$ are the Gaussian filter and the image, respectively, and $\star$ stands for the convolution operator.
    
    \item [b.] The Devernay Sub-Pixel Correction
    
    Edge points are attained with pixel precision through the Canny Algorithm, yet sub-pixel accuracy is desired in our system. This is where the Devernay Correction \cite{devernay1995subpixel} comes as an improvement of the former. Suppose $B$ is a pixel point obtained with the Canny Algorithm (a local gradient maximum in the corresponding normal direction $\mathbf{n}$). The gradients of two neighbouring points of $B$ along $\mathbf{n}$ can be approximated through linear interpolation. Then, the sub-pixel position of the edge point is refined to the maximum of a 1D quadratic interpolation (along $\mathbf{n}$) of all these three points.
    
    \item [c.] Contour Points Chaining
    
    The above independent points can be grouped to form several chains \cite{rafael2017implementation}, among which the one contains the most points is selected as the contour of the inner octagon, assuming a dense point detection (otherwise, the contour may be broken up into several pieces).
\end{itemize}

\subsection{Edge Line Fitting} \label{subsection: edge}
Given the set of 2D points $\mathbf{X}\in \mathbb{R}^{N\times 2}$ on the stop sign contour from above, eight lines are to be fitted, with each estimating an edge of the octagon.

The edges are estimated one at a time, by finding the line with the most "support points". A contour point "supports" a line when the point-to-line distance is within a closeness threshold (in this work, the threshold is fixed to 0.5 px to ensure a sub-pixel accurate estimation). This line selection is done by running a RANSAC algorithm over pairs of contour points (Algorithm \ref{alg: edge}). 

\begin{algorithm}[t] 
\SetAlgoLined
\For{$k\in \{1, ..., K\}$}{
Randomly sample two contour points;\\
Fit a line through them;\\
Find and count the support points of the line above;\\
Record the support points whenever the count is the highest so far;
}
Find the (ordinary) least square line of the recorded support points.\\
Determine the two end points of the line segment from the extrema of the recorded support points.
\caption{Estimating Edge $i$}
\label{alg: edge}
\end{algorithm} 

With $p\in [0, 1)$ probability that the final least square line is the most supported line, the number of iterations $K$ can be calculated as \cite{hartley2004multiple}:
\begin{equation}
    K = \frac{\log_2(1-p)}{\log_2(1-w^2)},
\end{equation}
where $w=\frac{1}{(8-i)^2} ,i\in \{0, ..., 7\}$ is the chance of selecting a point from the support set, assuming that there are few outliers in $\mathbf{X}$ off the contour. Determining $p$ involves considering a tradeoff between the estimating accuracy and the running time. Since currently the accuracy matters more than a high speed, we set $p$ as close to $1$ as possible (e.g. $0.999$). Notice here the number of iterations is bounded by ${N\choose 2}$ by iterating over all pairs.

Once an edge has been fitted, all the support points are removed from the original set, so that the same Algorithm \ref{alg: edge} can be repeated to find a different edge.

\subsection{Line Refinement with Perpendicular Search}
Primitive points from section \ref{subsection: contour} can stray away from the contour due to misleading color segmentation in the HSV space caused by poor illumination. Line refinement is therefore necessary to adjust the shifted lines back to the true octagon edges through a local search.

The magnitude of an (RGB) image gradient reveals the position of the octagon contour, where the color changes most abruptly in the neighborhood. Starting from an estimated line of section \ref{subsection: edge}, we search for the local maximal gradient along the perpendicular direction of the line, and rotate the line to align with it (Algorithm \ref{alg: refine}).

\begin{algorithm}[htbp] 
\SetAlgoLined
Sample $S$ points evenly from the line;\\
\For{each point}{
Search for the closest maximal gradient;\\
Record the refined point;
}
Find the (ordinary) least square line of the $S$ refined points.
\caption{Refining Line $i$}
\label{alg: refine}
\end{algorithm} 

In the case of under-/over-exposure, the gradient peak of an edge may disappear. Thus, a refinement boundary is introduced to cope with this. With the assumption that the refinement should be subtle, a line is only updated when more than half of the $S$ refined points are within the boundary. The refinement boundary is parameterized with a percentage $B\in [0, 1]$ (which is specified in each experiment later), meaning that a refined point is at most $100B$ percent of the white border width away from the original position along the perpendicular direction. To estimate the white border width of a stop sign, we multiply the length of an estimated line segment from Algorithm \ref{alg: edge} with the width-to-length ratio of an inner polygon edge documented in the traffic sign regulation.

\subsection{Estimation of Intersection Points} \label{subsection: intersection}
There are, at most, ${8\choose 2}=28$ intersection points from the eight estimated lines, while only eight of them are valid octagon corners. A simple yet empirically effective approach would be to select the intersection points close (say, within 10-pixel distance) to the end points of the line segments. After this, any estimation with a corner count not equal to eight is dropped.

Sorting the corner points in, for example, a clock-wise order (with respect to the geometric center of the points) helps obtain the 2D-3D correspondence. Further, the first corner can always be set as the one to the upper-left of the character "S", for stop signs generally stand upright. 

To further verify the validity of the corner points, the Hausdorff distance method in \cite{huttenlocher1990hausdorff} is adopted to guarantee that the final eight-point set forms an octagon under an affine transformation, which is, though two degrees-of-freedom less than the actual projective transformation, the best we can do without getting into the chicken-or-egg dilemma.

\subsection{Calibration with Planar Objects}
The pinhole camera model \cite{szeliski2010vision} is used.
A 3D point $(X, Y, Z)$ in the world coordinates is projected into an image as a pixel $(u, v)$. The following equation relates $(u,v)$ to $(X,Y,Z)$:
\begin{equation} \label{equ:camera_model}
s\left[\begin{array}{l}
u \\
v \\
1
\end{array}\right]=\mathbf{A}\left[\begin{array}{llll}
\mathbf{r}_{1} & \mathbf{r}_{2} & \mathbf{r}_{3} & \mathbf{t}
\end{array}\right]\left[\begin{array}{l}
X \\
Y \\
Z \\
1
\end{array}\right], \\
\end{equation}
where $s$ is an unknown scalar and $\mathbf{A} \in \mathbb{R}^{3 \times 3}$ is the camera intrinsic matrix (assuming skewless constrain). $\mathbf{r}_{1}, \mathbf{r}_{2}, \mathbf{r}_{3}$ are the columns of the rotation matrix, and $\mathbf{t}$ is the translation vector. They are often called the camera extrinsic parameters, for their function of transforming a world coordinate point to the camera coordinate system.

If the 3D coordinate system is fixed on a planar object like a chessboard or a stop sign, the $Z$ component in Equ. \ref{equ:camera_model} is always zero, which leads to a simplified equation:
\begin{equation}
s\left[\begin{array}{l}
u \\
v \\
1
\end{array}\right]=\mathbf{A}\left[\begin{array}{lll}
\mathbf{r}_{1} & \mathbf{r}_{2}  & \mathbf{t}
\end{array}\right]\left[\begin{array}{l}
X \\
Y \\
1
\end{array}\right]. \\
\end{equation}
We use $\mathbf{M}=(X,Y,1)^T$ and $\mathbf{m}=(u,v,1)^T$ to represent the position of a point on the planar object and  its position in the image, respectively. Their relationship is defined by a homography matrix $\mathbf{H}$:
\begin{equation} \label{equ:homography}
s \mathbf{m}=\mathbf{H M} \, \text{ with } \, \mathbf{H}=\mathbf{A}\left[\begin{array}{lll}
\mathbf{r}_{1} & \mathbf{r}_{2}  & \mathbf{t}
\end{array}\right]
\end{equation}
As is clear, $\mathbf{H}$ is a $3 \times 3$ matrix up to a scale factor. 

In \cite{hartley1998error}\cite{hartley1997defense}, the homography matrix $\mathbf{H}$ can be computed using $\mathrm{N} \geq 4$ points. In our system, 8 corner points of a stop sign are detected.

Let's represent $\mathbf{H}$ by $\mathbf{H}=\left[\begin{array}{lll}\mathbf{h}_{1} & \mathbf{h}_{2} & \mathbf{h}_{3}\end{array}\right]$. From Equ. \ref{equ:homography}, we can get:
\begin{equation}
\left[\begin{array}{lll}
\mathbf{h}_{1} & \mathbf{h}_{2} & \mathbf{h}_{3}
\end{array}\right]=\lambda \mathbf{A}\left[\begin{array}{lll}
\mathbf{r}_{1} & \mathbf{r}_{2} & \mathbf{t}
\end{array}\right]  \, \text{ with } \,  \lambda = \frac{1}{s}
\end{equation}
Based on the knowledge that $\mathbf{r}_{1}, \mathbf{r}_{2}$ are orthonormal, we get the two constraints needed for computing the intrinsic matrix A:
\begin{equation} \label{equ:constraints}
\begin{split}
\mathbf{h}_{1}^{T} \mathbf{A}^{-T} \mathbf{A}^{-1} \mathbf{h}_{2}&=0 \\
\mathbf{h}_{1}^{T} \mathbf{A}^{-T} \mathbf{A}^{-1} \mathbf{h}_{1}&=\mathbf{h}_{2}^{T} \mathbf{A}^{-T} \mathbf{A}^{-1} \mathbf{h}_{2}
\end{split}
\end{equation}
These are the two constraints needed for computing the intrinsic matrix $\mathbf{A}$. 

The methods of solving the camera calibration problem and obtaining the closed-form solution are thoroughly described in \cite{zhang2000calibration}. We refer interested readers to that paper for more information.

In our system, assuming that the principal point is fixed at the center of the image plane, only the focal lengths $f_x$ and $f_y$ are computed and iteratively updated as shown in the next section. For chessboard calibration, we also enforce this assumption for the purpose of comparison.

The closed-form solution can be refined through maximum likelihood inference: The original calibration is used as the initial guess for the nonlinear optimization problem, which can be solved with the Levenberg-Marquardt Algorithm as implemented in \cite{more1978levenberg}. Currently, this part is not included in our system.

\subsection{Temporal Updates of Calibration with a Kalman Filter}
The objective is to integrate intrinsic parameters over time. It is reasonable to assume that the estimation results are influenced by Gaussian noise, thus we use a Kalman filter with varying noise covariance matrices.

The prediction step is:
\begin{equation}
x_{t+1}=\mathbf{F}x_t + \mathbf{G}u_t + \epsilon,
\end{equation}
where $x_t, x_{t+1}$ are vectors of intrinsic parameters estimated at different times.
As there is no systematic manipulation over the intrinsic parameters, we can leave out the term $\mathbf{G}u_t$. If we believe that the intrinsic parameters are constant within a short period of estimation (which is a fundamental assumption of our research), $\mathbf{F}$ becomes the identity matrix. 

The update step is:
\begin{equation}
p_t=\bar{\mathbf{H}}x_t + \delta,
\end{equation}
which performs a refinement step of the parameters $x_t$ at time $t$." For this, $\bar{\mathbf{H}}$ is again assumed to be an identity matrix.

The process noise $\epsilon$ and the measurement noise $\delta$ are Gaussian random vectors with zero means. Their covariances are denoted by (diagonal) matrices $\mathbf{Q}=\mathbb{E}(\epsilon \epsilon^T), \mathbf{R}=\mathbb{E}(\delta \delta^T)$, respectively. $\mathbf{Q}$ specifies how much we think the intrinsic parameters change over time, which should be low for a short estimating period. $\mathbf{R}$ determines how much noise there is in one estimation. Intuitively, this measurement noise ought to decrease from a high value as the process goes, because the more images are used, a more accurate calibration is expected. 
\section{EXPERIMENTS} \label{section: experiments}
\subsection{Description of the Experimental Autonomous Vehicle}
\begin{figure}[H]
\centering
\includegraphics[angle=0,height=5.5cm,width=7cm]{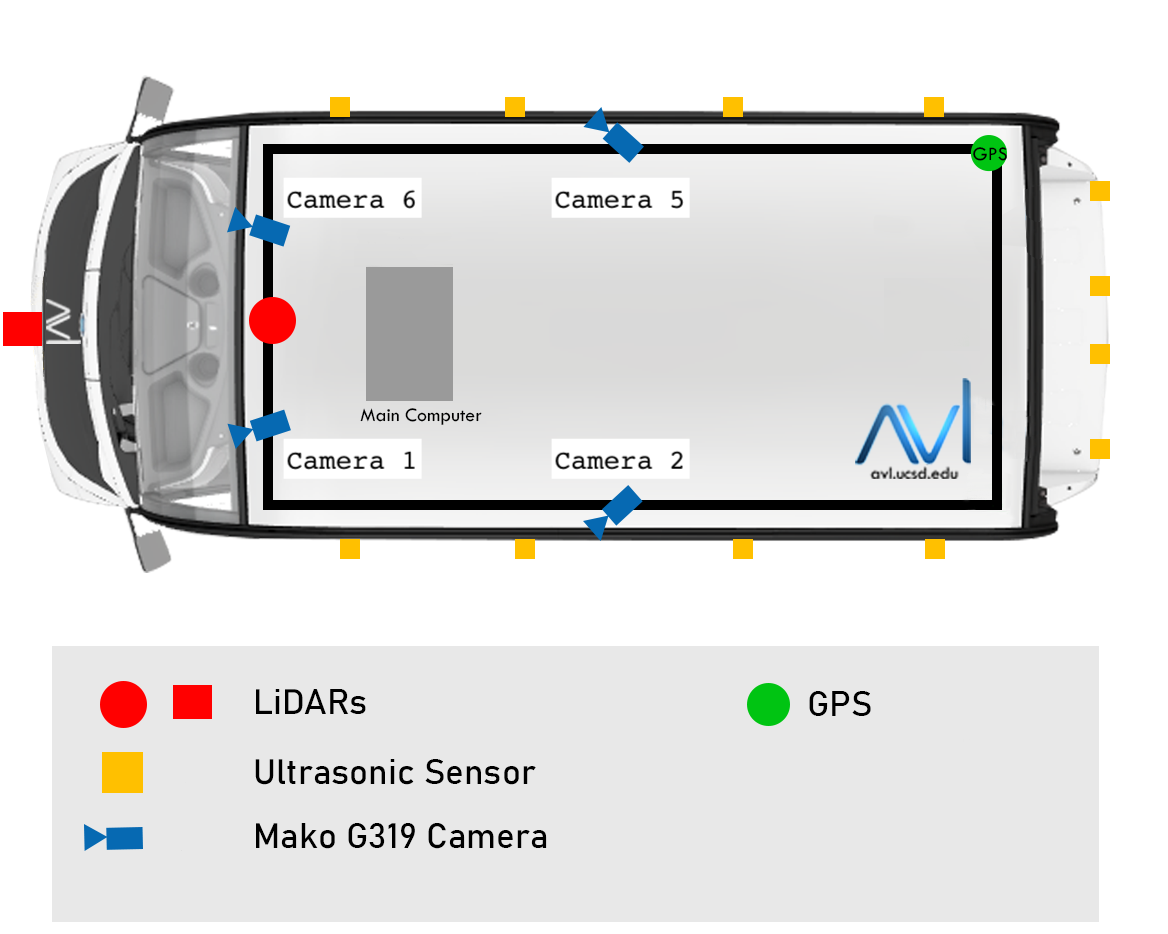}
\caption{Vehicle Sensor Configuration} \label{Vehicle}
\end{figure}
Our experiment data was collected by one of our experimental autonomous vehicles. The vehicle is equipped with four cameras. The cameras are set up as two on the front and one on each side (Fig. \ref{Vehicle}). Image streams of the cameras, data of LiDARs, and the vehicle positions were recorded for experiments by driving along multiple routes around the UC San Diego campus \cite{labPaper}. The camera video was streamed at approximately 12Hz. The camera data is used for our urban scenario experiments\footnote{More information can be found on our website: \href{http://avl.ucsd.edu/?page_id=592}{http://avl.ucsd.edu/}.}. 

\subsection{Datasets}
The experiments were done using datasets collected by the Autonomous Vehicle Lab (AVL). The image frames were stored in PNG formats: 

    
    
\textit{\textbf{StopSign:}}
    Only camera 5 and 6 were used and each detected 1507 and 1330 stop sign candidates using the Mask R-CNN, respectively. The ground-truth parameters (GT) were obtained from chessboard calibration beforehand.

\subsection{Results}
\label{section: results}
For scale-free comparison, the relative error $\epsilon_f$ of each parameter $f\in \{f_x, f_y\}$ is computed:
\begin{equation}
    \epsilon_f = \frac{f-f^{GT}}{f^{GT}},
\end{equation}
where $f^{GT}\in \{f_x^{GT}, f_y^{GT}\}$ is the ground-truth intrinsic parameter.

\hspace{-0.45cm}\textbf{\textit{ StopSign}}

\begin{figure}[htbp]
	\centering
	\includegraphics[angle=0,height=3cm,width=9cm]{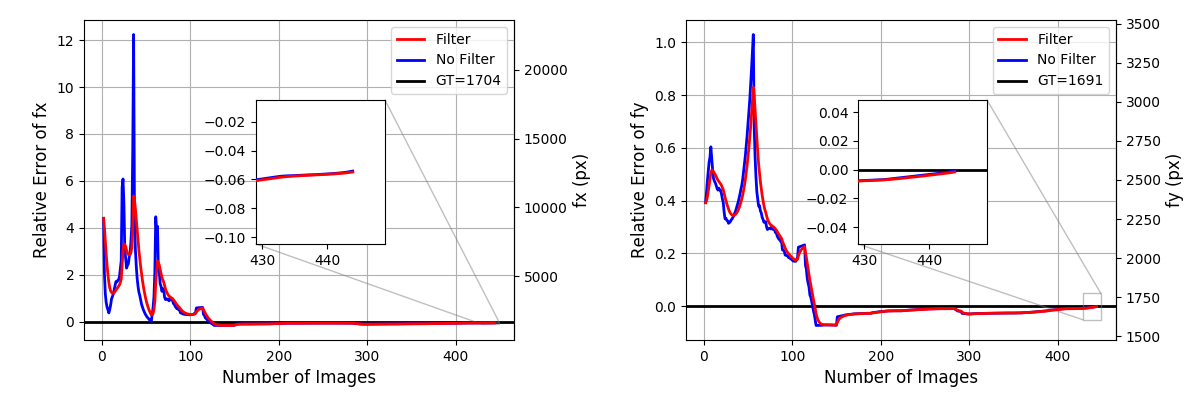}\\
	\caption{Calibration Results of StopSign-Camera5 Data Using Kalman Filters}
	\label{fig:v2_filter}
\end{figure}

\begin{figure}[htbp]
	\centering
	\includegraphics[angle=0,height=3cm,width=9cm]{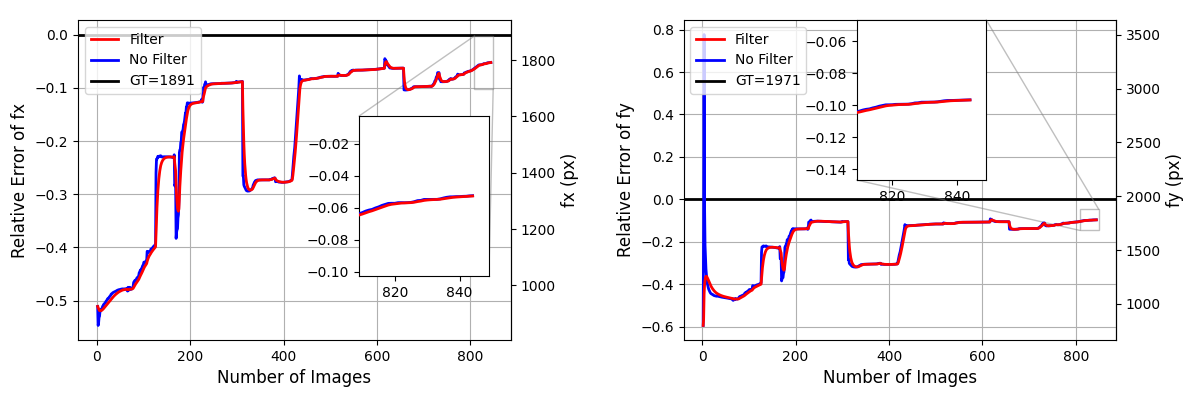}\\
	\caption{Calibration Results of StopSign-Camera6 Data Using Kalman Filters}
	\label{fig:v2_filter_appendix}
\end{figure}

\begin{figure}[htbp]
\centering
\includegraphics[angle=0,height=4.5cm,width=7cm]{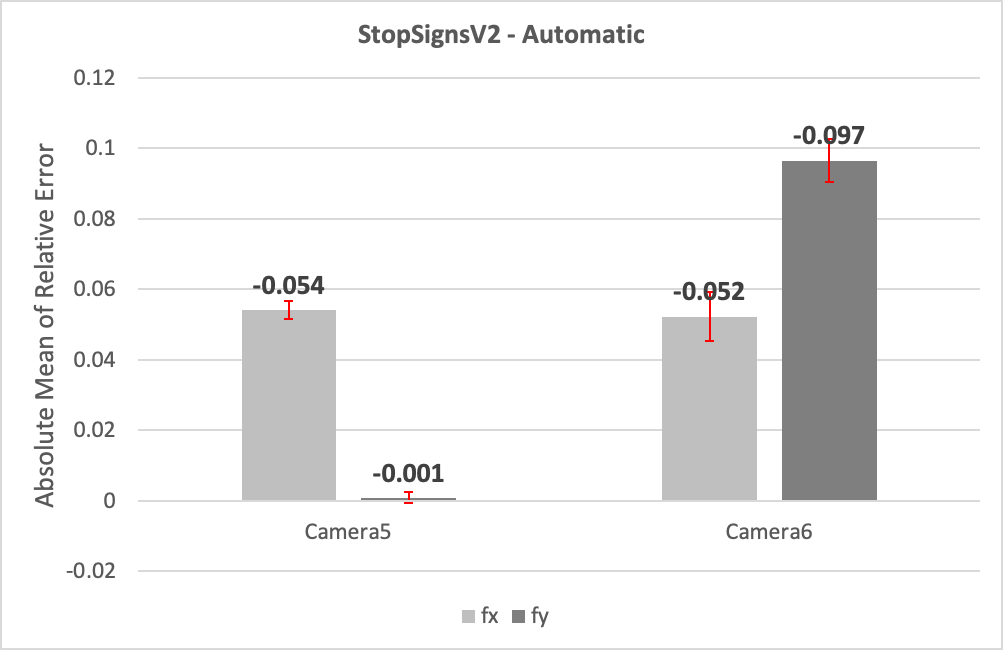}
\caption{Means and Standard Deviations of Relative Errors of StopSign Calibration without Kalman Filters} \label{fig:v2-calib-auto}
\end{figure}

For all of the two stop sign candidate sets, the refinement boundary percentages were: $B = 0.0$, i.e. no refinement. (Experiments showed that the line refinement was helpful when the detected stop signs were fewer. Due to the page limit, we omitted these edge cases because stops signs are readily found and well-distributed in urban scenarios as discussed in Section \ref{subsection: distriubtion}.) The system selected 444 (Camera5) and 844 (Camera6) good stop signs ready for calibration. System results with Kalman Filters are shown in Fig. \ref{fig:v2_filter} below and Fig. \ref{fig:v2_filter_appendix}. The summary of pure calibration is Fig. \ref{fig:v2-calib-auto}.


\subsection{Discussions}
Our approach converges to the correct parameters as shown in 
Fig. \ref{fig:v2_filter}. A 5\% relative error is achieved for \textit{StopSign}-Camera5 (Fig. \ref{fig:v2_filter}), which has over 400 stop signs for calibration. If we consider the 1-2\% variation of the chessboard calibration (verified by experiments), the above error can be as low as 3\% (and at most 7\% in the worst case). This is just 1\% higher than that of the chessboard method.



\subsection{Characterizing the Error} \label{subsection: char_error} 
In this section, we present to the readers intuitively the effect of the relative error using a simple example of distance measurement. Suppose the principal point of a camera is fixed at the center of the image plane. A 3D point $P=(X,Y,Z)^T$ in the camera coordinates is projected to a pixel $p=(u,v)^T$ in the image plane. The pinhole camera model gives the relation:
\begin{equation}
Z=f_x\cdot \frac{X}{u} \quad \text{and} \quad Z=f_y\cdot \frac{Y}{v}
\end{equation}
where $f_x, f_y$ are the focal lengths estimated using our system. If an algorithm measures the object-to-camera distance $Z$ from images captured by a camera calibrated using our system, the measurement error of $Z$ (assuming the algorithm itself doesn't introduce any error) is equal to the relative error of the estimated focal lengths $\epsilon_{f_x}, \epsilon_{f_y}$. Take \textit{StopSign}-Camera5 as an example (which achieved a 5\% relative error), an object estimated at 50m away will be located between 47.62 and 52.63m ($\frac{50m}{1\pm 5\%}$). This error range is acceptable for reacting safely in urban areas with a 25mph (11.175m/s) speed limit: emergency braking experiments from 25mph to 0mph indicate that on downhill roads, 27.7m (4.4s) are needed during emergency stops using our autonomous driving platform. 

\subsection{Distribution of Stop Signs} \label{subsection: distriubtion} 
To verify the feasibility of leveraging stop signs for automatic calibration, we characterize the distribution of stop signs in an urban scenario and estimate the expected number of detections per unit distance. 
While uptime is another unit that could be used for normalization, in heavy traffic scenarios the distribution may be biased if the vehicle remains at a standstill.

\begin{figure}[htbp]
\centering
\includegraphics[angle=0,height=7cm,width=8.5cm]{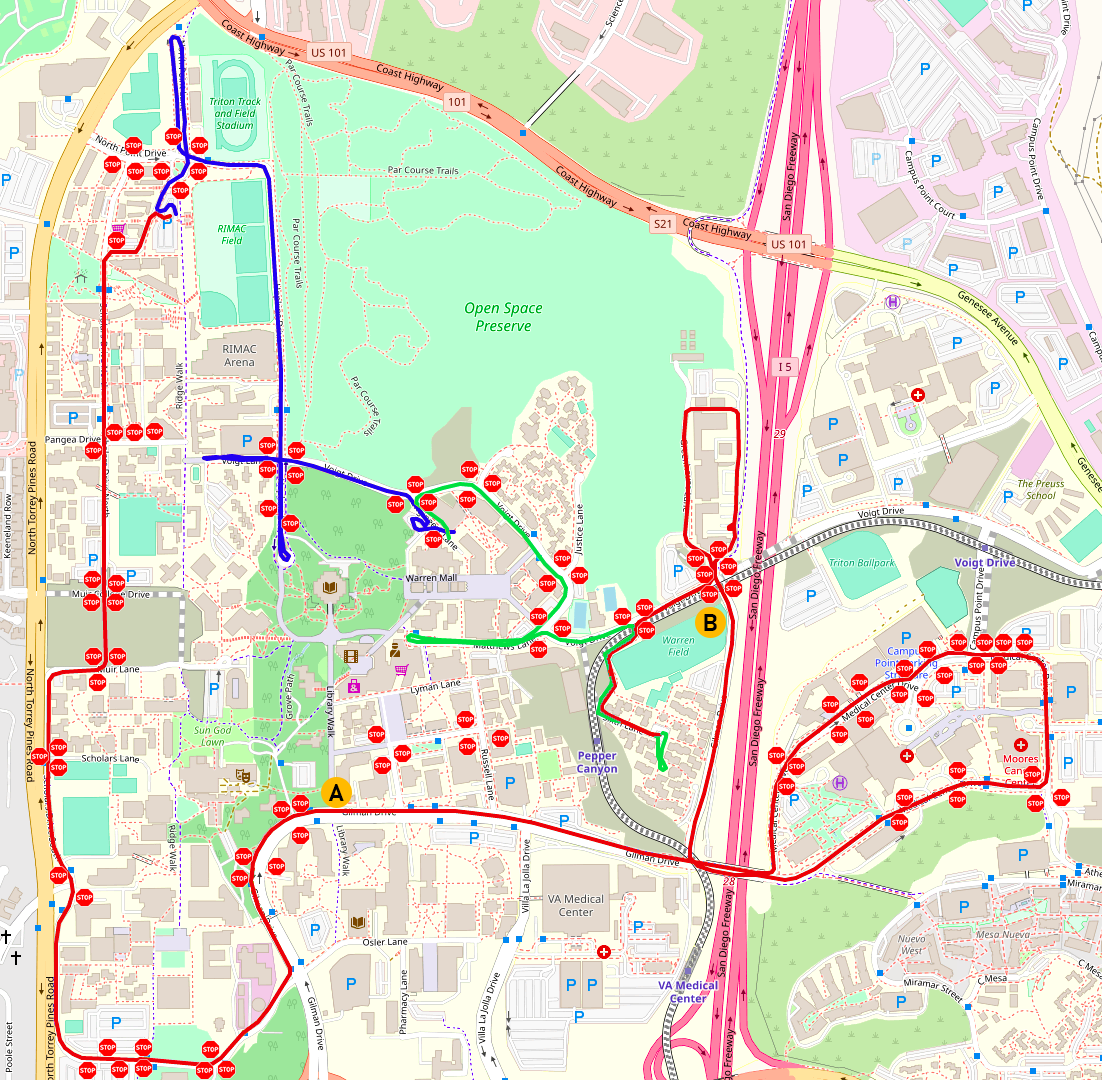}
\caption{A demonstration of a campus loop and the stop sign distribution. The blue, green, and red trajectories correspond to the three routes the vehicle drove in order. Calibrations were done at the intersections of consecutive routes.}
\label{fig:stop-sign-distribution}
\end{figure}

To quantify the expected number of the detections per unit distance, we compute the expected number of detections per stop sign and the number of stopsigns per unit distance. For the UCSD campus, an average of 33.96(detections/stop sign) were determined by using our dataset from the passenger-front camera (at 12 FPS). A campus wide experiment shown in Fig. \ref{fig:stop-sign-distribution} consists of 43 stop sign encounters over a distance of 12.02km / 7.47mi (blue, green and red trajectories combined). This is equivalent to $\frac{33.96\times 43}{12.02} = 121.5$(detections/km) or $\frac{33.96\times 43}{7.47} = 195.5$(detections/mi), which is sufficient for automatic calibration as previously shown in Section \ref{section: results}.


In the worst case, the vehicle may have to drive for 0.9 miles as shown between point A and point B in Fig. \ref{fig:stop-sign-distribution} to find a stop sign; however, stop signs are just one example of traffic signs (e.g. LISA traffic sign dataset has 47 U.S. sign types). Our system can be readily extended to use many of these references for calibration that can be found between points A and B. Even though the passenger-right camera was used for this experiment, by leveraging additional traffic signs that are oriented at various angles and sides, we can address the shortcomings with certain traffic signs that cannot be observed by a camera.

\subsection{Intrinsic Variation} \label{subsection: variation} 
To study the variation of the calibration parameters during regular driving, we divided the full trajectory of Fig. \ref{fig:stop-sign-distribution} into three routes and performed chessboard calibration at both ends of each route (four calibrations in total). The calibration results are shown in TABLE \ref{tab:recalibration-results}. (The relative variation is computed by $\frac{\max -\min}{\text{average}}$.)

\begin{table}[h]
    \centering
    \begin{tabular}{c|c|c}
    \hline
         Calibrations (Distance) &  $f_x$ (pixel) & $f_y$ (pixel) \\
        \hline
    Reference (0km/0mi)    &   1810.4 & 1840.1 \\
        \hline
    Stop1 (2.63km/1.63mi)    &   1870.1 & 1906.3 \\
        \hline
    Stop2 (1.85km/1.15mi)    &   1918.7 & 1939.0  \\
        \hline
    Stop3 (7.54km/4.69mi)    &   1935.2 & 1965.8 \\
        \hline
    Relative Variation  &   6.6256\% & 6.5715\% \\
        \hline
    \end{tabular}
    \caption{The calibration results at three stops.}
    \label{tab:recalibration-results}
\end{table}


Additional experiments were also performed by driving over 1, 3 and 10 speedbumps with calibrations in between; however, the calibration variation was within the calibration error range. This implies that calibration degradation is influenced more by distance and not necessarily by single events, motivating the auto-calibration system. Furthermore, our system can achieve an error smaller than the relative variations of focal lengths, which demonstrates the effectiveness. 

\subsection{Non-Standard and Occluded Stop Signs} \label{subsection: Non-standard}   
Here we analyze the robustness of our system to ill-conditioned stop signs detected in real scenes.
\begin{figure}[htbp]
	\begin{subfigure}{0.24\textwidth}
	\includegraphics[angle=0,height=3.6cm,width=4.2cm]{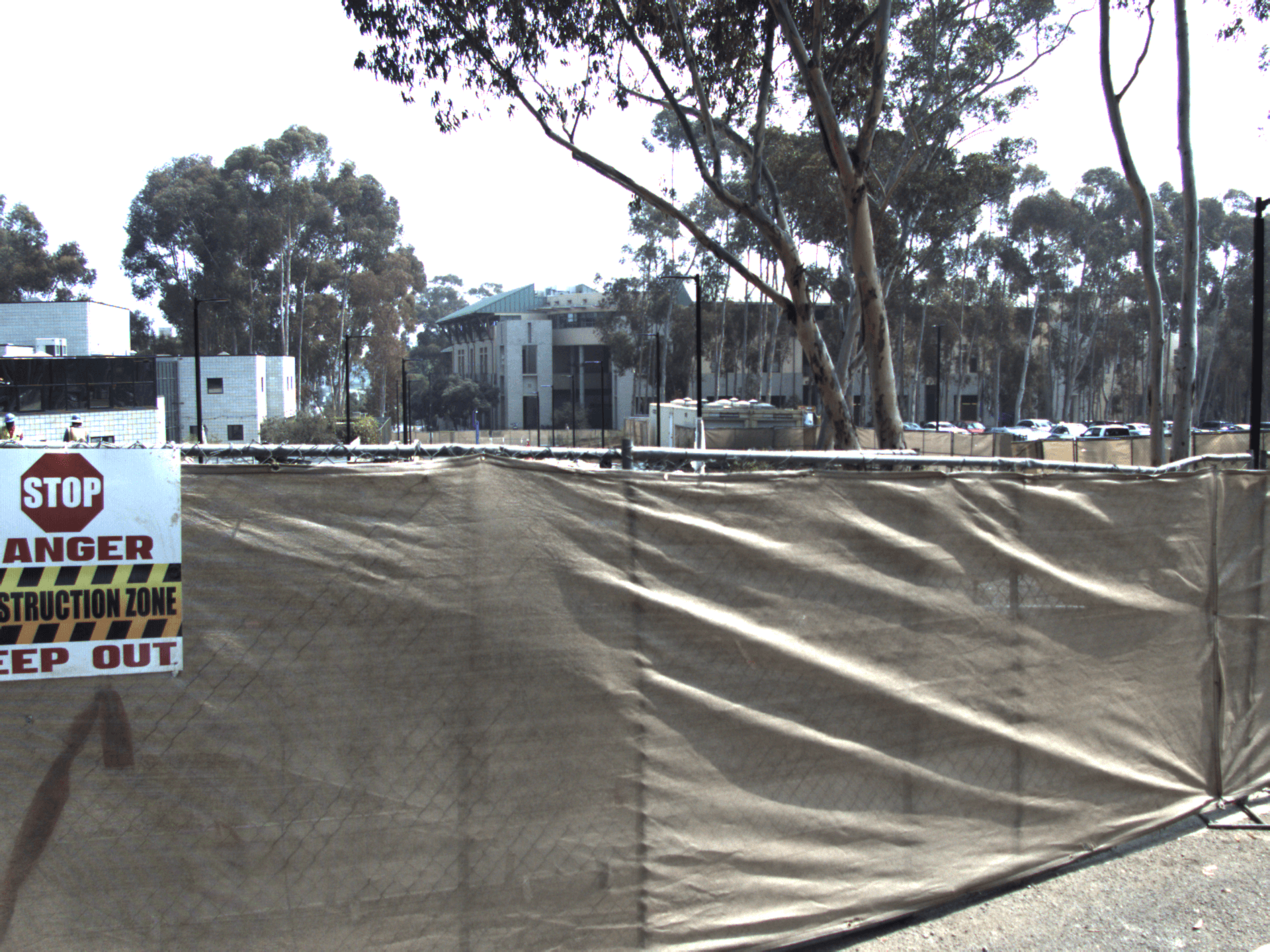}
	\end{subfigure}
	\begin{subfigure}{0.24\textwidth}
	\includegraphics[angle=0,height=3.6cm,width=4.2cm]{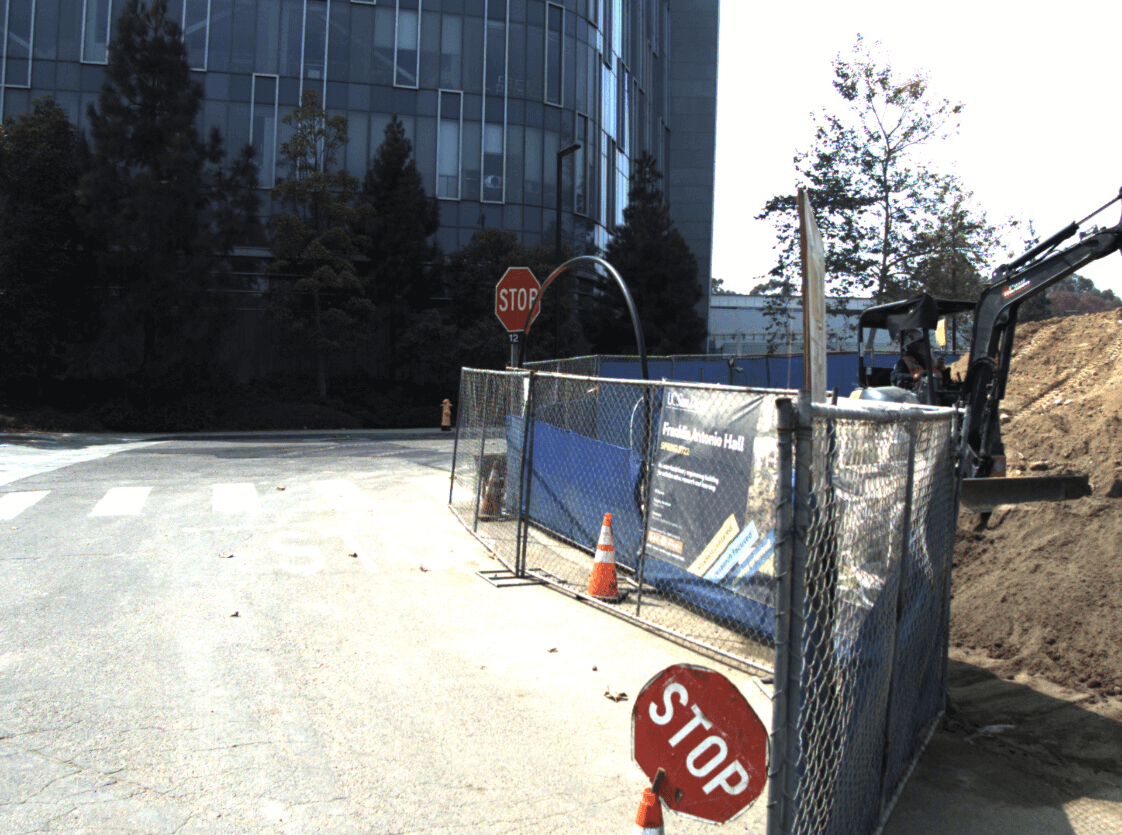}
	\end{subfigure}
	\caption{Two stop sign examples that are removed by our approach. The stop sign in the left image has a non-standard geometry and the stop signs in the right image are occluded by other objects.}
	 \label{fig:non_standard}
\end{figure}
The system automatically removed the non-standard stop sign in the left image because of a detection failure using the Mask R-CNN. The occluded stop signs in the right image were also removed because the chain which contains the most points did not form an octagon under an affine transformation (as is previously described in section \ref{subsection: intersection}).


\section{CONCLUSION}
A system for online calibration of the cameras has been presented. The system is well suited for urban environments with a rich set of traffic signs. The system uses recursive estimation of camera parameters and demonstrates robustness in detection and updating. The use of natural features for calibration compared to artificial patterns such as a chess board allow for deployment on vehicles in natural environments. Experiments in a campus environment are used to demonstrate the performance.  
Experiments show that there were 1-2\% variations in the chessboard calibration, which means the relative errors of our system may be smaller as the Kalman Filter converges. 
The system framework has the advantage of being easily adapted to using other traffic signs or landmarks. This allows us to create an auto-calibration toolbox, which will greatly benefit the whole autonomous driving community.

For future work, we will replace the naive HSV color space segmentation in the contour points detection with a more robust method (e.g. semantic segmentation), to improve the contour points detection under undesirable lighting conditions. Also, we will introduce the Levenberg-Marquardt algorithm to refine the calibration results and track more parameters. Experiments will be carried out to compare the refined results with the current results. 

\section*{ACKNOWLEDGMENT}
Many thanks to Hengyuan Zhang for chessboard calibration, Andres Gutierrez for data pre-processing and all AVL lab members for the insightful discussions.

\newpage
{\small
\bibliographystyle{./IEEEtran}
\bibliography{IEEEcitation}
}
\end{document}